\documentclass[runningheads]{llncs}

\usepackage[T1]{fontenc}

\usepackage{times}
\usepackage{soul}
\usepackage{url}
\usepackage{graphicx}
\usepackage{amsmath}
\usepackage{booktabs}

\usepackage{multicol}
\usepackage{multirow}
\usepackage{color}
\usepackage{xcolor}
\usepackage{booktabs}
\usepackage{graphicx}
\usepackage{nicefrac}
\usepackage{amsmath}
\usepackage{array}
\newcolumntype{P}[1]{>{\centering\arraybackslash}p{#1}}

\usepackage{graphicx}
\usepackage{fancyhdr}
\usepackage{url}
\usepackage{color}
\usepackage{marvosym}

\usepackage{float}

\usepackage{caption}
\usepackage{subcaption}

\usepackage{xfp} 

\usepackage{longtable}
\usepackage{booktabs}
\usepackage{amsfonts}
\usepackage{algorithm}
\usepackage{algorithmicx}%
\usepackage{algpseudocode}%
\usepackage{textcomp}
\usepackage{xfp}
\usepackage{multicol}
\usepackage{multirow}
\usepackage{color}
\usepackage{xcolor}
\usepackage{booktabs}
\usepackage{graphicx}
\usepackage{nicefrac}
\usepackage{amsmath}
\usepackage{caption}
\usepackage{subcaption}
\usepackage{array}
\newcolumntype{P}[1]{>{\centering\arraybackslash}p{#1}}
\usepackage{graphicx}
\usepackage{fancyhdr}
\usepackage{amssymb}
\usepackage{color}
\usepackage{marvosym}
\usepackage{float}
\usepackage{hyperref}
\usepackage{color}
\usepackage{xcolor}
\usepackage{xspace}
\newcommand*\BitAnd{\mathbin{\&}}

\begin{document}

\title{Hilbert Curve Based Molecular Sequence Analysis}







\author{Sarwan Ali$^{1*}$\and
Tamkanat E Ali$^{2*}$\and 
Imdad Ullah Khan \inst{2}\and 
Murray Patterson \inst{1}
}
\authorrunning{S. Ali et al.}

\institute{$^1$Department of Computer Science, Georgia State University, Atlanta, Georgia, USA
\\
$^2$Department of Computer Science, Lahore University of Management Sciences, Lahore, Pakistan
\\
\email{sali85@student.gsu.edu, 20100159@lums.edu.pk, imdad.khan@lums.edu.pk, mpatterson30@gsu.edu}
\\
$^*$Equal Contribution
}

\maketitle

\begin{abstract}
Accurate molecular sequence analysis is a key task in the field of bioinformatics. To apply molecular sequence classification algorithms, we first need to generate the appropriate representations of the sequences. Traditional numeric sequence representation techniques are mostly based on sequence alignment that faces limitations in the form of lack of accuracy. Although several alignment-free techniques have also been introduced, their tabular data form results in low performance when used with Deep Learning (DL) models compared to the competitive performance observed in the case of image-based data. To find a solution to this problem and to make Deep Learning (DL) models function to their maximum potential while capturing the important spatial information in the sequence data, we propose a universal Hibert curve-based Chaos Game Representation (CGR) method. This method is a transformative function that involves a novel Alphabetic index mapping technique used in constructing Hilbert curve-based image representation from molecular sequences. Our method can be globally applied to any type of molecular sequence data. The Hilbert curve-based image representations can be used as input to sophisticated vision DL models for sequence classification. The proposed method shows promising results as it outperforms current state-of-the-art methods by achieving a high accuracy of $94.5$\% and an F1 score of $93.9\%$ when tested with the CNN model on the lung cancer dataset. This approach opens up a new horizon for exploring molecular sequence analysis using image classification methods.

\end{abstract}

\section{Introduction}\label{sec_intro}
With the rapid increase in biological sequence data, there is a growing challenge of effective data analysis, mining, and visualization. Molecular Sequence analysis plays a crucial role in disease detection, drug discovery, etc. To analyze molecular sequences such as DNA, Proteins, etc using ML and DL-based algorithms there is a need to generate appropriate representations of their basic units including nucleotides and amino acids.

The numeric vector-based embeddings~\cite{iuchi2021representation} have been widely used to form molecular sequence representations and have shown promising results when used with ML classifiers but when Deep Learning classification models such as Neural networks (NN) are applied to the numeric vectors they perform suboptimally as NN models struggle with tabular data~\cite{shwartz2022tabular} due to issues like feature sparsity, varying scales, and the lack of spatial correlations. As a solution image-based molecular sequence embeddings~\cite{akbari2022walkim} were introduced that show improved performance when used with DL-based classification models. One of the most widely used image-based representation methods is based on Chaos game representation (CGR)~\cite{lochel2021chaos} which is a groundbreaking invention in the field of graphical bioinformatics and has gained increasing popularity as it has proved to be successful in the effective image-based encoding of biological sequence features for the accurate application of Deep learning algorithms as it effectively maps a sequence to 2-dimensional space. It has also proved to be a powerful tool for alignment-free sequence comparison. The CGR is based on the concept of the Markov chain and it allows a unique representation of a sequence. On the other hand Frequency matrix representation (FCGR) which is an extension of (CGR), takes sequences of different lengths as input and transforms them into equal-sized images or matrices.

Despite the promising performance of Deep Learning (DL) classification models when used with these image-based sequence representations, most existing methods face a major limitation in accurately capturing all the important information preserved in the data while keeping the accuracy intact as they only consider sequence information regardless of spatial information~\cite{zhang2021epishilbert}. To provide a solution to this challenge Hilbert curve-based image representations~\cite{yin2018image} were introduced to represent molecular sequences in a manner that not only helped perform significantly well in DL-based classification tasks but also was successful in capturing the important spatial information in the sequence data. A Hilbert curve~\cite{sagan1994hilbert} is a continuous fractal space-filling curve that maps 1-dimensional sequences to a two-dimensional plane. Hilbert curve is considered better than other space-filling curves such as z-order and Peano curves because it has much better spatial cohesion~\cite{ghosh2015hilbert}, spatial
continuity, and self-similarity~\cite{cardona2016self}. Hilbert curve has been widely used for visualizing genomic data~\cite{nusrat2019tasks}. These current Hilbert curve-based representation methods fail to provide a universal solution as their encoding is specific to the type of molecular sequence used. To overcome this limitation we propose a universal Hibert curve-based Chaos Game Representation (CGR) method that involves a unique Alphabetic index mapping technique used in constructing Hilbert curve-based image representations from molecular sequences. This technique can be globally applied to any type of molecular sequence. These Hilbert curve-based representations are further used as input to DL models for sequence classification. 
Our study presents the following key contributions:
\begin{itemize}
    \item We introduce a novel universal Hibert curve-based Chaos Game Representation (CGR) method for converting molecular sequences into image representations. It involves a novel Alphabetic index mapping technique used in constructing Hilbert curve-based image representation from molecular sequences enabling the application of advanced vision deep learning models for sequence classification tasks.
    \item Our approach is rigorously tested on various molecular sequence datasets, showcasing significant improvements in predictive performance compared to traditional methods.
    \item The proposed methodologies are not limited to the molecular sequences but are also applicable to other sequence datasets from the NLP domain, illustrating the versatility and broad impact of our work.
\end{itemize}


\section{Related Work}\label{sec_RW}
Several methods have been used to generate numeric vector-based embeddings to represent molecular sequences such as Feature Engineering, Natural Language Processing (NLP), Neural Networks, and Kernel-based methods. Feature Engineering-based embedding generation methods is a well-explored field with significant discoveries including PWM2Vec~\cite{ali2022pwm2vec} that involves weight assignment strategy for each amino acid in a $k$-mer based on its position for generating numeric representation, some prominent peptide descriptors include amino acid composition (AAC)~\cite{fu2020acep}, Composition of k-spaced amino acid group pairs (CKSAAGP)~\cite{usman2020afp}, pseudo amino acid composition (PAAC)~\cite{beltran2024multitoxpred} and physicochemical features (PHYC). In another research~\cite{pang2021identifying}, a combination of such descriptors is used for anti-CoV peptides classification.
Neural Network-based molecular sequence embeddings have been the target of extensive research. Autoencoder has been used by multiple researchers for generating molecular sequence representations~\cite{hadipour2022deep}. 
Natural Language Processing (NLP) has gained popularity in the field of bioinformatics. NLP-based methods have been widely used for molecular sequence embedding generation, some recent works include SeqVec~\cite{heinzinger2019modeling}, PRoBERTa~\cite{nambiar2020transforming}, MSA-transformer~\cite{rao2021msa}, and ESM2~\cite{susanty2024classifying}. 
Other than numeric representations, image-based representations are also used to represent molecular sequences which are further used as input to DL-based image classification algorithms. Most widely used Image-based representations of molecular sequences involve Chaos game representation strategy such as Spike2CGR~\cite{murad2023spike2cgr} which converts spike sequences to graphical (image) representation, FCGR~\cite{halder2024improved} converts DNA sequences to images, SARS-CoV-2 lineages, strains and recombinants are classified using CGR based representations~\cite{thind2023using}.
Hilbert curve-based image representations~\cite{yin2018image} have also been used to represent molecular sequences.

\section{Proposed Approach}\label{sec_PA}
In this section, we present the details of our proposed universal Hilbert curve-based Chaos Game Representation (CGR) method for transforming biological sequences into image representations suitable for deep learning classification. The core idea is to map one-dimensional biological sequences onto a two-dimensional plane using a Hilbert curve, which preserves spatial locality and captures essential features of the sequences.

The Hilbert curve is a continuous fractal space-filling curve~\cite{sagan1994hilbert}, which is used for mapping data from a one-dimensional space to a two-dimensional space that preserves locality. This essentially means that the points that are close in the one-dimensional space remain close in the two-dimensional mapping. This property is particularly advantageous for encoding molecular sequences because it maintains the proximity of sequence elements in the image representation.

Mathematically, the Hilbert curve of order $p$ in $N$ dimensions can be defined recursively. The total number of points $\Theta$ on the curve is given by $\Theta = 2^{p*N}$, where $p$ is the number of iterations (or the order of the curve), and $N$ is the number of dimensions. Since we are generating the 2-dimensional image, the value for $N$ is taken as $2$. To generate the images of standard size (i.e. $64 \times 64$), we use $p=6$ to get $2^6 \times 2^6 = 64 \times 64$ dimensional image for any given sequence. Note that when $p$ represents the number of iterations in constructing the Hilbert curve, the dimensions of the resulting image are determined by the expression $2^p \times 2^p$. Each increment in $p$ leads to an exponential increase in the size of the image, allowing the curve to fill more space and represent more complex data or patterns effectively.   

\begin{figure}[h!]
    \centering
    \includegraphics[scale=0.05]{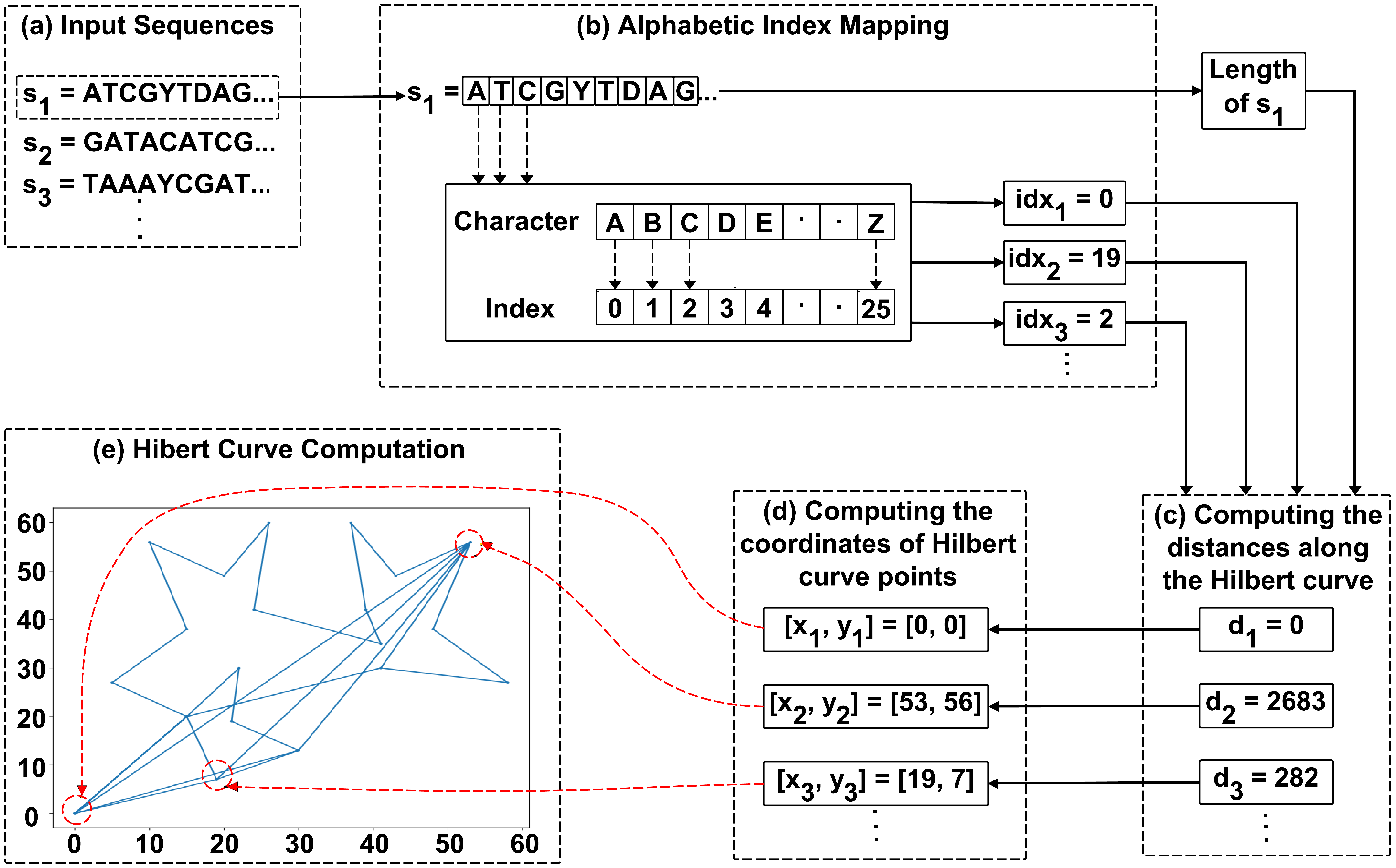}
    \caption{Flow diagram for the proposed method.}
    \label{flow_diagram}
\end{figure}

\paragraph{Hilbert Curve Computation}
Given a set of biological sequences $S = \{s_1, s_2, \ldots, s_n\}$, where each sequence $s_i$ consists of characters from the alphabet set $A$, we generate Hilbert curve-based image representations using the Algorithm~\ref{algo_hilbert_curve}.
The goal is to map each character in $s_i$ to a unique point on the Hilbert curve, resulting in an image representation of the sequence.
The flow diagram is presented in Figure~\ref{flow_diagram}, which gives a bigger picture of the steps involved in our proposed approach.

\paragraph{Alphabetic Index Mapping:}
We define an alphabetic index mapping function 
IndexMapping(c) in Algorithm~\ref{algo_Index_mapping} that assigns a unique integer index $I$ to each character $c$ in the alphabet $A$:
\begin{equation}
\scriptsize
    I = IndexMapping(c), c \in A
\end{equation}
This mapping ensures that each character is represented by a distinct numerical value (see Step (b) in Figure~\ref{flow_diagram}).

\paragraph{Distance Calculation:}
For each character $c$ in the sequence $s_i$, we calculate a normalized distance $D$ along the Hilbert curve as shown in line number 8 of Algorithm~\ref{algo_hilbert_curve}  (also see Step (c) in Figure~\ref{flow_diagram}):
\begin{equation}
\scriptsize
 D = \frac{I}{L}*\Theta
\end{equation}
where $L$ is the length of the sequence $s_i$, and $\Theta$ is the total number of points on the Hilbert curve. This equation scales the index $I$ proportionally to the sequence length and maps it onto the range of the Hilbert curve.

\begin{remark}
    The combination of index mapping and distance calculation ensures a bijective relationship between sequence characters and points on the Hilbert curve within the resolution defined by $p$. This means each character maps to a unique point, preventing information loss.
\end{remark}

\paragraph{Computing Hilbert Curve Coordinates:}
The next step involves converting the distance $D$ into two-dimensional coordinates $(x,y)$ on the Hilbert curve using Algorithm~\ref{algo_Points}  (also see Step (d) in Figure~\ref{flow_diagram}). This is achieved through a series of transformations involving binary representations and Gray codes.

\paragraph{Binary Representation:}
We convert the distance $D$ into a binary number with $n=p \times N$ bits as shown in line number 2 and 3 of Algorithm~\ref{algo_Points}:

\begin{equation}
\scriptsize
    Bits = Binary (D) = b_{n-1} b_{n-2} \ldots b_0
\end{equation}

\paragraph{Bit Interleaving:}
We separate the bits into even and odd indices to form the coordinate components as shown in lines number 7 and 8 of Algorithm~\ref{algo_Points}:

\begin{equation}
\scriptsize
    EvenIdxBits = b_{n-1} b_{n-3} \cdots
\end{equation}

\begin{equation}
\scriptsize
    OddIdxBits = b_{n-2} b_{n-4} \cdots
\end{equation}
These binary sequences are then converted back to decimal form to obtain preliminary coordinate values as shown in line numbers 14 and 16 of Algorithm~\ref{algo_Points}:

\begin{equation}
\scriptsize
    x_{raw} = Decimal(EvenIdxBits)
\end{equation}

\begin{equation}
\scriptsize
    y_{raw} = Decimal(OddIdxBits)
\end{equation}

\paragraph{Gray Code Transformation:}
We apply the Gray code transformation to the coordinate components to minimize bit changes between successive values, preserving spatial locality. We apply this transformation by calling the function '\textsc{GenerateGrayCode()}' in line number 18 of Algorithm~\ref{algo_Points}. The function \textsc{GenerateGrayCode()} is defined in Algorithm~\ref{algo_Gray_Code} and the following equations:  

\begin{equation}
\scriptsize
    x_{gray} = x_{raw} \oplus (x_{raw} \gg 1)
\end{equation}

\begin{equation}
\scriptsize
    y_{gray} = y_{raw} \oplus (y_{raw} \gg 1)
\end{equation}
where $\oplus$ denotes the bitwise XOR operation and $\gg$ denotes a bitwise right shift.

Algorithm~\ref{algo_Gray_Code} applies Global $np$-fold Gray code~\cite{skilling2004programming} to all bits of each integer representing distance along the Hibert Curve at once yielding a result that is quite close to the desired spatial coordinates.

\paragraph{Coordinate Refinement:}
During the process of Gray Code transformation large number of exchanges and inversions are conducted resulting in an over-transformed length. The excess transformation is undone using the function '\textsc{Refining()}' in line 19 of Algorithm~\ref{algo_Points} that works according to Algorithm~\ref{algo_clean_up} that refines the Gray code to obtain the final Hilbert curve coordinates $(x,y)$.
This involves inverting the Gray code transformation:
\begin{equation}
\scriptsize
    x = InverseGrayCode(x_{gray})
\end{equation}

\begin{equation}
\scriptsize
    y = InverseGrayCode(y_{gray})
\end{equation}
The inverse Gray code transformation is performed iteratively:

\begin{equation}
    \begin{aligned}
    \scriptsize
        &x = x_{gray} \\
        &For \ i = n-2 \ to \ 0: \\
        &x_i = x_i \oplus x_{i+1}
    \end{aligned}
\end{equation}
The same above process applies to $y_{gray}$ to obtain $y$.
The equations above conduct the exchanges and inversions needed to transform Gray Code into the desired coordinates. The refined output $(x,y)$ gives us the Hilbert Curve point coordinates.

\begin{algorithm}[h!]
\caption{Hilbert Curve Computation}
\label{algo_hilbert_curve}
\begin{algorithmic}[1]
\scriptsize
\Statex \textbf{Input: }\texttt{Set of Biological Sequences ($S$), Number of Iterations $p$, Number of Dimensions $N$, Alphabets ($A$)}
\Statex \textbf{Output: }\texttt{Hilbert Curve  Coordinates($H$)}
\Function{ComputeHilbertCurve}{$S$,$p$,$N$}
\State H $\gets$ [ ]
\State $\Theta$ = $2^{(p * N)}$
\Comment{Compute number of unit hypercubes using p = 6 and N = 2}

\For{\texttt{ $s_{1}$ in S\hspace{0.2cm}}}
    \State L $\gets$ Length($s_{1}$)
    
    \For{\texttt{ $char$ in $s_{1}$\hspace{0.2cm}}}
    \State Index $\gets$ \Call{IndexMapping}{$char$, $A$}
    \State D $\gets$ $\dfrac{Index}{L}$ * $\Theta$ \Comment{Compute distances along Hilbert curve}
    \State Coordinates $\gets$ PointfromDistance(D) \Comment{Single Hilbert curve point coordinates} 
    \State $H$.append(Coordinates)
    \EndFor
\EndFor
\State \Return $H$ 
\EndFunction
\end{algorithmic}
\end{algorithm}

\begin{algorithm}[h!]
\caption{Alphabetic Index Mapping}
\label{algo_Index_mapping}
\begin{algorithmic}[1]
\scriptsize
\Statex \textbf{Input: }\texttt{Biological Sequence single character($c$), Alphabets (A)}
\Statex \textbf{Output: }\texttt{Index($I$)}

\Function{IndexMapping}{$c$, $A$}
\State
\For{\texttt{ $a$ in $A$\hspace{0.2cm}}}

    \If{a = $c$}
    \State $I$ = $A$.index($c$) 
    \Comment{Find the alphabetic index of single sequence character}
    \EndIf
\EndFor
\State \Return $I$ 
\EndFunction
\end{algorithmic}
\end{algorithm}

\begin{algorithm}[h!]
\caption{Distance to Hilbert Curve Point}
\label{algo_Points}
\begin{algorithmic}[1]
\scriptsize
\Statex \textbf{Input: }\texttt{Distance along Hilbert Curve ($D$), Number of Iterations ($p$)}
\Statex \textbf{Output: }\texttt{Point($\phi$)}
\Function{PointfromDistance}{$D$}
\State NumBits = $N*p$ \Comment{Number of bits}
\State Bits = Binary($D$)
\Comment{Distance to Binary string representation of $N$*$p$ bits}
\State OddIdxBits $\gets$ [ ]
\State EvenIdxBits $\gets$ [ ]
\State Components $\gets$ [ ]

\For{i in range(0, LenBits)}
\If{i \% 2 = 0}
    \State EvenIdxBits.append(Bits[i])
    
    \ElsIf{i \% 2 = 1}
    \State OddIdxBits.append(Bits[i])
    \EndIf
\EndFor
\State EvenIdxDeci = Decimal(EvenIdxBits)
\State Components.append(EvenIdxDeci)
\State OddIdxDeci = Decimal(OddIdxBits)
\State Components.append(OddIdxDeci)
\State GrayCode $\gets$ \Call{GenerateGrayCode}{Components, $n$}
\State $\phi$ $\gets$ \Call{Refining}{GrayCode, $p$}
\State \Return $\phi$ 
\EndFunction
\end{algorithmic}
\end{algorithm}

\begin{algorithm}[h!]
\caption{Gray Code Computation}
\label{algo_Gray_Code}
\begin{algorithmic}[1]
\scriptsize
\Statex \textbf{Input: }\texttt{Components ($C$),$n$}
\Statex \textbf{Output: }\texttt{Gray Code($\gamma$)}
\Function{GenerateGrayCode}{$C$}
\State $r$ = $C$[$n$-1] $\gg$ 1 \Comment{Right shift}

\For{i in range($n$-1, 0, -1)}
    \State $C$[i] $\gets$ $C$[i] $\oplus$ $C$[i-1] \Comment{Bitwise XOR between two consecutive bits}
    \EndFor
\State $C$[0] $\gets$ $C$[0] $\oplus$ $r$
\State $\gamma$ $\gets$ $C$
\State \Return $\gamma$ 
\EndFunction
\end{algorithmic}
\end{algorithm}

\begin{algorithm}[h!]
\caption{Refining}
\label{algo_clean_up}
\begin{algorithmic}[1]
\scriptsize
\Statex \textbf{Input: }\texttt{Gray Code Integers($g$), Number of Iterations ($p$)}
\Statex \textbf{Output:} \texttt{Point($\beta$)}
\Function{Refining}{$g$}
       \State q = 2
       \State z = 2 $\ll$ ($p$-1)

        \While{q != z}
         \State w = q - 1
         
            \For{i in range(n-1, -1, -1)}
                \If{$g$[i] $\BitAnd$ q}
                    \Comment{Check if the specific bit that is ON in q is also ON in $g$[i]}
                    \State $g$[0] = $g$[0] $\oplus$ w \Comment{Invert low bits of $g$[0]}
                    \Else
                    \State t = ($g$[0] $\oplus$ $g$[i]) $\BitAnd$ w 
                    \State $g$[0] = $g$[0] $\oplus$ t\Comment{Exchange low bits of $g$[0] and $g$[i]}
                    \State $g$[i] = $g$[i] $\oplus$ t
                    \EndIf
                    
            \EndFor
            \State q = q $\ll$ 1
        \EndWhile
\State $\beta$ $\gets$ $g$
\State \Return $\beta$ 
\EndFunction
\end{algorithmic}
\end{algorithm}

\subsection{Mathematical Justification}

\paragraph{Gray Code Utilization:}
Gray codes are employed to minimize the Hamming distance between successive integer values, which translates to minimal movement between adjacent points on the Hilbert curve. This is crucial for maintaining the continuity and smoothness of the curve, and it enhances the preservation of sequential information in the image representation.

\paragraph{Refinement Process:}
The refinement step corrects the over-transformation caused by the global Gray code application. By performing inversions and exchanges in a backward pass through the higher-order bits, we ensure that the final coordinates accurately reflect the intended mapping.

\subsection{Collision Guarantee}
The Hilbert curve is a space-filling curve, and thus, as $p \rightarrow \infty$, the curve becomes denser, filling all points in the space. For a finite $p$, the guarantee is that the curve will cover all unit hypercubes without any overlap within the resolution limit set by $\Theta = 2^{p \times N}$.

\subsection{Upper Bound on Uniqueness}
Since the Hilbert curve is injective for any given resolution, there is no loss of information within the resolution constraints. As long as the number of unit hypercubes exceeds the sequence length (i.e., $\Theta > L$), the curve guarantees a unique mapping of characters to curve points. If $\Theta \leq L$, multiple characters may map to the same spatial location, leading to potential loss of information.



\section{Experimental Setup}\label{sec_ES}
To compare results for proposed method with existing approaches, we divided the baselines into vector-based and image-based methods. For vector-based, we use one hot encoding OHE~\cite{kuzmin2020machine}, Spike2Vec~\cite{ali2021spike2vec}, Minimizer spectrum~\cite{girotto2016metaprob}, Spaced $k$-mers spectrum~\cite{singh2017gakco}, PWM2Vec~\cite{ali2022pwm2vec}, Wasserstein Distance Guided Representation Learning (WDGRL)~\cite{shen2018wasserstein}, Auto-Encoder~\cite{xie2016unsupervised}, and pre-trained LLM known as SeqVec~\cite{heinzinger2019modeling}. For image-based baselines, we use Frequency Chaos Game Representation (FCGR)~\cite{lochel2020deep}, Spike2CGR~\cite{murad2023spike2cgr}, and Random Chaos Game Representation (RANDOMCGR)~\cite{murad2023new}.

In this study, we employed two categories of classifiers: vector-based and image-based. 
In the vector-based setting, we use the standard nearest neighbor classifier, which is well-known and traditionally used in the literature for such tasks. 
For the image-based classifiers, we use various deep-learning architectures, including 1, 2, and 3 layers of Convolutional Neural Network (CNN), Visual Geometry Group (VGG19), ResNet50, EfficientNet, and DenseNet.
For the parameters of the DL models, we selected values for batch size, epochs, and learning rate as $64$, $10$, and $0.003$, respectively. For the optimizer, we use ADAM. For hyperparameter tuning, we use a standard cross-validation setting.
To evaluate the performance of the classifiers, we use average accuracy, precision, recall, weighted F1, Macro F1, ROC-AUC, and training runtime.



The dataset used in this study consists of peptide sequences classified based on their anticancer activity against breast and lung cancer cell lines~\cite{Grisoni2019}. These anti-cancer peptides (ACP) are categorized into four groups: "very active," "moderately active," "experimental inactive," and "virtual inactive," reflecting their varying levels of anticancer effectiveness. This makes it a 4-class classification problem.
The breast cancer dataset contains $949$ peptide sequences, while the lung cancer dataset includes $901$ sequences. The peptide lengths vary across categories, with the minimum sequence length being $5$ and the maximum $38$. On average, sequences range from $14.5$ to $20.7$ amino acids long, depending on their activity level. This distribution ensures a diverse set of peptide lengths and activity profiles, facilitating a robust evaluation of the proposed method.
For data splitting, we use standard $80$ percent for training and $20$ percent for testing (i.e. held-out set). From the training data, $10$ percent is used for validation.
All experiments are conducted on an Intel i5 processor (2.40 GHz) and 32 GB of RAM. The operating system used is Windows 10, and the code is implemented using Python. 

\section{Results And Discussion}\label{sec_RD}
In this section, we present and discuss the results of our classification experiments on both the breast cancer dataset and the second dataset.

The results for the breast cancer ACPs are reported in Table~\ref{tbl_breast_cancer_avg_data_results}. The vector-based methods, which are used as traditional baselines (other than Auto-Encoder and WDGRL), show sub-optimal performance overall. For the Auto-Encoder, we can observe a higher predictive performance, achieving an accuracy of $83.2\%$, a weighted F1 score of $80.4\%$, and a ROC-AUC of $64.5\%$. This highlights the utility of feature learning in extracting meaningful representations from the sequences. 
WDGRL, a domain adaptation approach, provides comparatively higher performance with an accuracy of $79.4\%$. However, it fell short in terms of F1 macro and ROC-AUC metrics, indicating the potential misalignment between the source and target domains in this scenario.

\begin{table}[h!]
    \centering
    \resizebox{0.8\textwidth}{!}{
         \begin{tabular}{p{1.9cm}p{1.8cm}p{3.3cm}p{1cm}p{1.3cm}p{1.3cm}p{1.6cm}cp{1.6cm} | p{1.8cm}}
    \toprule
    \multirow{2}{1.1cm}{Method} & \multirow{2}{1.1cm}{Algorithm} & \multirow{2}{1.1cm}{ML/DL Model}  & \multirow{2}{*}{Acc. $\uparrow$} & \multirow{2}{*}{Prec. $\uparrow$} & \multirow{2}{*}{Recall $\uparrow$} & \multirow{2}{1.6cm}{F1 weigh. $\uparrow$} & \multirow{2}{1.4cm}{F1 Macro $\uparrow$} & \multirow{2}{1.5cm}{ROC- AUC $\uparrow$} & Train. runtime (sec.) $\downarrow$ \\
    \midrule \midrule	
        \multirow{11}{2.5cm}{Vector Based} 
        & \multirow{1}{1.2cm}{OHE} & -
         & 0.609 & 0.853 & 0.609 & 0.676 & 0.395 & 0.678 & 0.069   \\
        \cmidrule{3-10}  
         & \multirow{1}{1.2cm}{Spike2Vec} & -
        & 0.241 & 0.298 & 0.241 & 0.212 & 0.200 & 0.550 & 0.133   \\
        \cmidrule{3-10} 
         & \multirow{1}{1.2cm}{Minimizer}  & -
        & 0.577 & 0.807 & 0.577 & 0.635 & 0.332 & 0.616 & 0.149   \\

        \cmidrule{3-10}  
         & \multirow{1}{2.5cm}{Spaced k-mer}  & -
        & 0.276 & 0.460 & 0.276 & 0.253 & 0.216 & 0.559 & 1.036   \\
        \cmidrule{3-10} 
         & \multirow{1}{1.2cm}{PWM2Vec} & -
         & 0.199 & 0.808 & 0.199 & 0.221 & 0.190 & 0.541 & 0.618     \\

        \cmidrule{3-10}        
         & \multirow{1}{1.9cm}{WDGRL} & -
        & 0.794 & 0.715 & 0.794 & 0.730 & 0.270 & 0.518 & \textbf{0.016}   \\
        \cmidrule{3-10} 
         & \multirow{1}{2.5cm}{Auto-Encoder} & -
        & 0.832 & 0.802 & 0.832 & 0.804 & 0.431 & 0.645 & 0.067     \\
        \cmidrule{3-10} 
         & \multirow{1}{1.5cm}{SeqVec} & -
        & 0.674 & 0.819 & 0.674 & 0.725 & 0.389 & 0.651 & 22.253  \\
        \midrule
         \multirow{28}{2.5cm}{Image Based} 
         & \multirow{7}{2.5cm}{FCGR} 
          & 1 Layer CNN & 0.863 & 0.831 &  0.863 & 0.844 & 0.490 & 0.677 & 5410.357 \\
          & & 3 Layer CNN & 0.800 & 0.640 &  0.800 & 0.711 & 0.222 & 0.500 & 52147.851 \\
          & & 4 Layer CNN & 0.831 & 0.735 & 0.831 & 0.779 & 0.329 & 0.586 & 56873.749 \\
          & & VGG19 (pre-trained) & 0.803 &  0.684 & 0.803 & 0.720 & 0.243 & 0.509 &  51234.241 \\
          & & RESNET50 (pre-trained) & 0.800 & 0.642 & 0.800 & 0.712 & 0.222 & 0.501 & 49715.758 \\
          & & Efficient Net & 0.089 & 0.008 & 0.089 & 0.014 & 0.041 & 0.500 & 8731.614 \\
          & & Dense Net & 0.116 & 0.013 & 0.116 & 0.024 & 0.052 & 0.500 & 11482.259 \\
          \cmidrule{2-10} 
          & \multirow{7}{2.5cm}{Spike2CGR} 
          & 1 Layer CNN & 0.783 & 0.613  & 0.783 & 0.687 & 0.219  & 0.500  & 6547.979  \\
          & & 3 Layer CNN & 0.783 & 0.612  & 0.783  & 0.687 & 0.219  & 0.500  & 55419.449  \\
          & & 4 Layer CNN & 0.783 & 0.612  & 0.783  & 0.687 & 0.219  & 0.500  & 56482.458  \\
          & & VGG19 (pre-trained) & 0.765 & 0.650  & 0.765  & 0.650 & 0.200  & 0.500  & 49851.852  \\
          & & RESNET50 (pre-trained) & 0.770 & 0.559  & 0.770  & 0.654 & 0.198  & 0.500  & 50179.716  \\
          & & Efficient Net & 0.085 & 0.005  & 0.085  & 0.009 & 0.008  & 0.500  & 9812.984  \\
          & & Dense Net & 0.116 & 0.011  & 0.116  & 0.022 & 0.050  & 0.500  & 10248.154  \\
          \cmidrule{2-10}
          & \multirow{7}{2.5cm}{RandomCGR} 
          & 1 Layer CNN & 0.792 & 0.638 & 0.792 &  0.707 & 0.221 &  0.497 & 4982.864 \\
          & & 3 Layer CNN & 0.800 & 0.640 &  0.800 & 0.711 & 0.222 & 0.500 & 53214.341 \\
          & & 4 Layer CNN & 0.800 & 0.640 &  0.800 & 0.711 & 0.222 & 0.500 & 64128.387 \\
          & & VGG19 (pre-trained) & 0.800 & 0.640 &  0.800 & 0.711 & 0.222 & 0.500 & 53214.524 \\
          & & RESNET50 (pre-trained) & 0.800 & 0.640 &  0.800 & 0.711 & 0.222 & 0.500 & 55654.851 \\
          & & Efficient Net & 0.028 & 0.002 & 0.028 & 0.004 & 0.027 & 0.500 & 9547.759 \\ 
          & & Dense Net & 0.095 & 0.011 & 0.095 & 0.010 & 0.095 & 0.500 & 10247.751 \\ 
         \cmidrule{2-10}
          & \multirow{7}{2.5cm}{Ours} 
         & 1 Layer CNN & \textbf{0.895} & \textbf{0.869} & \textbf{0.895} & \textbf{0.881} & \textbf{0.521} & \textbf{0.725} & 2136.810 \\
         & & 3 Layer CNN & 0.842 & 0.834 & 0.842 & 0.838 & 0.429 & 0.672 & 13544.320 \\
         & & 4 Layer CNN & 0.874 & 0.861 & 0.874 & 0.867 & 0.476 & 0.705 & 13044.890 \\
         & & VGG19 (pre-trained) & 0.863 & 0.774 & 0.863 & 0.815 & 0.407 & 0.675 & 25081.410 \\
         & & RESNET50 (pre-trained) & 0.853 & 0.837 & 0.853 & 0.841 & 0.465 & 0.690 & 11202.436 \\
         & & Efficient Net & 0.084 & 0.007 & 0.084 & 0.013 & 0.039 & 0.500 & 3057.010 \\
         & & Dense Net & 0.800 & 0.640 & 0.800 & 0.711 & 0.222 & 0.500 & 2819.100 \\
    \bottomrule
  \end{tabular}  
    }
    \caption{Classification results for \textbf{Breast Cancer dataset}.}
    \label{tbl_breast_cancer_avg_data_results}
\end{table}

The image-based methods showed a wide range of performance across different CNN architectures. The proposed method consistently outperformed other image-based methods, achieving the highest accuracy of $89.5\%$, precision of $86.9\%$, recall of $89.5\%$, and an F1-weighted score of $88.1\%$ when using the 1-Layer CNN architecture. This indicates that our sequence-to-image transformation approach combined with a simple CNN model was effective in capturing the underlying patterns in the molecular sequence data.
Among other CNN architectures, 4-Layer CNN also provided competitive results, with an accuracy of $87.4\%$ and an F1-weighted score of $86.7\%$, while the 3-Layer CNN achieved slightly lower performance. Interestingly, EfficientNet shows poor performance with accuracies as low as $8.4\%$, suggesting that the deeper networks struggled to generalize well on the molecular image data.
In general, the CNN-based models, especially the simpler architectures like the 1-Layer CNN, excelled at the task of sequence classification, likely due to their ability to capture both local and global patterns in the image representation of the sequences. 

For the lung cancer dataset results, as shown in Table~\ref{tbl_lungs_cancer_avg_data_results}, the vector-based methods also present suboptimal performance for methods like PWM2Vec. The Auto-Encoder, however, delivered improved results, with an accuracy of $91.0\%$, a weighted F1 score of $90.8\%$, and a ROC-AUC of $77.1\%$, suggesting its effectiveness in capturing important sequence characteristics for lung cancer ACPs. The WDGRL, while providing an accuracy of $86.2\%$, exhibited lower scores in F1 macro and ROC-AUC, reinforcing the challenge of aligning the source and target domains effectively for this specific dataset.

\begin{table}[h!]
    \centering
   \resizebox{0.7\textwidth}{!}{
         \begin{tabular}{p{1.9cm}p{1.8cm}p{3.3cm}p{1cm}p{1.3cm}p{1.3cm}p{1.6cm}cp{1.6cm} | p{1.8cm}}
    \toprule
    \multirow{2}{1.1cm}{Method} & \multirow{2}{1.1cm}{Algorithm} & \multirow{2}{1.1cm}{ML/DL Model}  & \multirow{2}{*}{Acc. $\uparrow$} & \multirow{2}{*}{Prec. $\uparrow$} & \multirow{2}{*}{Recall $\uparrow$} & \multirow{2}{1.6cm}{F1 weigh. $\uparrow$} & \multirow{2}{1.4cm}{F1 Macro $\uparrow$} & \multirow{2}{1.5cm}{ROC- AUC $\uparrow$} & Train. runtime (sec.) $\downarrow$ \\
    \midrule \midrule	
        \multirow{9}{2.5cm}{Vector Based} 
        & \multirow{1}{1.2cm}{OHE} & -
         &  0.804 & 0.907 & 0.804 & 0.835 & 0.537 & 0.781 & 0.117   \\
        \cmidrule{3-10}  
         & \multirow{1}{1.2cm}{Spike2Vec} & -
        & 0.877 & 0.919 & 0.877 & 0.883 & 0.590 & 0.790 & 0.590   \\
        \cmidrule{3-10} 
         & \multirow{1}{1.2cm}{Minimizer}  & -
        & 0.858 & 0.835 & 0.858 & 0.840 & 0.455 & 0.681 & 0.837   \\

        \cmidrule{3-10}  
         & \multirow{1}{2.5cm}{Spaced k-mer}  & -
        & 0.883 & 0.871 & 0.883 & 0.862 & 0.530 & 0.699 & 21.594      \\
        \cmidrule{3-10} 
         & \multirow{1}{1.2cm}{PWM2Vec} & -
         & 0.452 & 0.842 & 0.452 & 0.511 & 0.335 & 0.614 & 0.931   \\

        \cmidrule{3-10}        
         & \multirow{1}{1.9cm}{WDGRL} & -
        & 0.862 & 0.820 & 0.862 & 0.822 & 0.360 & 0.583 & \textbf{0.050}   \\
        \cmidrule{3-10} 
         & \multirow{1}{2.5cm}{Auto-Encoder} & -
        & 0.910 & 0.908 & 0.910 & 0.906 & 0.602 & 0.771 & 0.090   \\
        \cmidrule{3-10} 
         & \multirow{1}{1.5cm}{SeqVec} & -
        & 0.886 & 0.882 & 0.886 & 0.878 & 0.604 & 0.761 & 33.326  \\
        \midrule
         \multirow{28}{2.5cm}{Image Based} 
         & \multirow{7}{2.5cm}{FCGR} 
          & 1 Layer CNN & 0.910 & 0.911 & 0.910 & 0.910 & 0.582 & 0.755 & 5023.028 \\
          & & 3 Layer CNN & 0.930 & 0.925 & 0.930 & 0.929 & \textbf{0.681} & \textbf{0.810} & 41247.742 \\
          & & 4 Layer CNN & 0.909 & 0.912 & 0.909 & 0.911 & 0.587 & 0.751 & 42215.749 \\
          & & VGG19 (pre-trained) & 0.921 & 0.919 & 0.921 & 0.918 & 0.600 & 0.776 & 59713.943 \\
          & & RESNET50 (pre-trained) & 0.915 & 0.918 & 0.915 & 0.914 & 0.598 & 0.777 & 49853.749 \\
          & & Efficient Net & 0.101 & 0.012 & 0.101 & 0.023 & 0.059 & 0.500 & 9024.137 \\
          & & Dense Net & 0.231 & 0.030 & 0.231 & 0.031 & 0.061 & 0.500 & 9851.749 \\
          \cmidrule{2-10} 
          & \multirow{7}{2.5cm}{Spike2CGR} 
          & 1 Layer CNN & 0.833 & 0.779  & 0.833 & 0.764 & 0.291  & 0.551  & 5987.149  \\
          & & 3 Layer CNN & 0.831 & 0.780  & 0.831  & 0.749 & 0.587  & 0.548  & 58745.217  \\
          & & 4 Layer CNN & 0.825 & 0.771  & 0.825  & 0.751 & 0.585  & 0.545  & 59412.743  \\
          & & VGG19 (pre-trained) & 0.805 & 0.852  & 0.805  & 0.851 & 0.573  & 0.544  & 50125.126  \\
          & & RESNET50 (pre-trained) & 0.837 & 0.799  & 0.837  & 0.843 & 0.555  & 0.541  & 51249.354  \\
          & & Efficient Net & 0.054 & 0.011  & 0.054  & 0.015 & 0.019  & 0.509  & 8712.258  \\
          & & Dense Net & 0.324 & 0.021  & 0.324  & 0.030 & 0.095  & 0.507  & 11423.017  \\
          \cmidrule{2-10}
          & \multirow{7}{2.5cm}{RandomCGR} 
          & 1 Layer CNN & 0.854 & 0.798 & 0.854 &  0.814 & 0.314 &  0.588 & 5024.749 \\
          & & 3 Layer CNN & 0.853 & 0.791 &  0.853 & 0.801 & 0.302 & 0.580 & 51249.149 \\
          & & 4 Layer CNN & 0.852 & 0.784 &  0.852 & 0.795 & 0.310 & 0.567 & 67418.249 \\
          & & VGG19 (pre-trained) & 0.892 & 0.714 &  0.892 & 0.769 & 0.297 & 0.524 & 60214.143 \\
          & & RESNET50 (pre-trained) & 0.890 & 0.701 &  0.890 & 0.755 & 0.294 & 0.532 & 51478.215 \\
          & & Efficient Net & 0.035 & 0.003 & 0.035 & 0.006 & 0.032 & 0.500 & 8745.149 \\ 
          & & Dense Net & 0.099 & 0.015 & 0.099 & 0.014 & 0.098 & 0.500 & 11427.137 \\ 
         \cmidrule{2-10}
          & \multirow{7}{2.5cm}{Ours} 
         & 1 Layer CNN & \textbf{0.945} & \textbf{0.938} & \textbf{0.945} & \textbf{0.939} & 0.664 & 0.791 & 1648.930 \\
         & & 3 Layer CNN & 0.906 & 0.915 & 0.906 & 0.908 & 0.563 & 0.758 & 16365.380 \\
         & & 4 Layer CNN & 0.912 & 0.909 & 0.912 & 0.909 & 0.534 & 0.729 & 15161.590 \\
         & & VGG19 (pre-trained) & 0.917 & 0.888 & 0.917 & 0.898 & 0.490 & 0.683 & 19264.990 \\
         & & RESNET50 (pre-trained) & 0.906 & 0.909 & 0.906 & 0.900 & 0.551 & 0.713 & 15041.838 \\
         & & Efficient Net & 0.061 & 0.876 & 0.061 & 0.027 & 0.031 & 0.503 & 5037.620 \\
         & & Dense Net & 0.873 & 0.762 & 0.873 & 0.814 & 0.233 & 0.500 & 3961.028 \\
 
    \bottomrule
  \end{tabular}  
    }
    \caption{Classification results for \textbf{Lungs Cancer dataset}.}
    \label{tbl_lungs_cancer_avg_data_results}
\end{table}

The image-based methods displayed a broader range of outcomes across different CNN models, similar to the breast cancer results. Our proposed method, employing the 1-Layer CNN, outperformed all other approaches with an accuracy of $94.5\%$, precision of $93.8\%$, recall of $94.5\%$, and an F1-weighted score of $93.9\%$. This demonstrates the strength of our sequence-to-image transformation technique in revealing patterns within the lung cancer molecular sequence data. The 3-Layer CNN and 4-Layer CNN also showed promising results, however, deeper architectures such as EfficientNet struggled again, showing a low accuracy of $6.1\%$. Overall, the simpler CNN models, particularly the 1-Layer CNN, performed exceptionally well, suggesting that less complex architectures are better suited for capturing informative features from molecular sequence images in the lung cancer dataset.
According to Occam's razor~\cite{walsh1979occam}, the simplest solution is often the best one. A 1-layer CNN may strike the optimal balance between model simplicity and learning capability for this particular task, where the sequence-to-image transformation provides enough meaningful features that don’t require deeper processing, leading to better generalization on unseen data.

\section{Conclusion}\label{sec_conclusion}
In this work, we proposed a novel approach for classifying molecular sequences by transforming them into images using the sequence-to-image transformation technique, called the Hilbert curve. 
The results demonstrate the potential of our method in capturing intricate patterns within molecular sequence data, particularly when coupled with CNN architectures. 
Future work could explore integrating advanced domain adaptation techniques to further enhance model generalization across different datasets, as well as experimenting with hybrid architectures that combine the strengths of vector-based and image-based approaches. 








\bibliographystyle{splncs04}
\bibliography{references}

\end{document}